\documentclass[conference]{IEEEtran}
\IEEEoverridecommandlockouts

\usepackage{cite}
\usepackage{amsmath,amssymb,amsfonts}
\usepackage{algorithmic}
\usepackage{graphicx}
\usepackage{textcomp}
\usepackage{xcolor}
\usepackage{lineno}
\usepackage[colorlinks=true, urlcolor=blue, linkcolor=red]{hyperref}
\usepackage{makecell}
\usepackage{placeins}

\def\BibTeX{{\rm B\kern-.05em{\sc i\kern-.025em b}\kern-.08em
    T\kern-.1667em\lower.7ex\hbox{E}\kern-.125emX}}
    
\begin{document}

\title{FoggyTrust: Robust Federated Learning with Hierarchical Trust Networks}

\author{
\IEEEauthorblockN{1\textsuperscript{st} Emmanuel Rassou}
\IEEEauthorblockA{\textit{School of Engineering and Applied Sciences (SEAS)} \\
\textit{Harvard}\\
Cambridge, MA, USA \\
emmanuel\_rassou@college.harvard.edu}
\and
\IEEEauthorblockN{2\textsuperscript{nd} Tomas Gonzalez}
\IEEEauthorblockA{\textit{School of Engineering and Applied Sciences (SEAS)} \\
\textit{Harvard}\\
Cambridge, MA, USA \\
tsgonzalez@college.harvard.edu}
}

\maketitle

\begin{abstract}
Byzantine-robust federated learning seeks to protect distributed model training from malicious or corrupted clients without requiring access to their private data. FLTrust addresses this challenge by introducing a trusted server-side root dataset that assigns trust scores to client updates for more robust aggregation. In this work, we propose \textsc{FoggyTrust}, a hierarchical extension of FLTrust that localizes trust computation to fog nodes, allowing the framework to better handle globally heterogeneous data while preserving robustness within locally homogeneous client groups. We further show that this two-level architecture can simultaneously address distribution mismatch in trust estimation and client drift across groups by combining local trust-based aggregation with heterogeneity-aware global optimizers such as FedAdam and SCAFFOLD. Across benchmark datasets, \textsc{FoggyTrust} achieves its strongest gains on more challenging heterogeneous settings, particularly on CIFAR-10 under Krum and Trim attacks, where it achieves an over 50\% improvement over FLTrust. We also test \textsc{FoggyTrust} in a real-world safari dataset to show the promise of  hierarchical trust networks for robust federated learning in socially impactful, safety-critical settings such as distributed wildlife monitoring.
\end{abstract}


\section{Introduction}
Federated Learning (FL) is an emerging distributed machine learning paradigm which enables learning from clients without exposing their data\cite{Konecny2016FLStrategies,FedAvg,FLTrust,Scaffold}. Specifically, FL uses a centralized service provider which maintains the global model. At each iteration of the learning algorithm, the service provider (i.e Google, Apple, etc) sends the global model to each client (i.e. smartphones, etc.), which then uses on-device data to make local updates to the model. The local updates are then sent back to the service provider, which uses an aggregation rule to fold all local updates into one global update. 

Since each client only sends their update to the parameters, it does not have to expose the data to the service provider, promising valuable advantages in privacy with various use cases\cite{Gboard, Biomedical}. However, the privacy is a double edged sword. Without access to the data used to make the local update, the service provider is unable to validate whether updates are correct, uncorrupted, and non-malicious. Thus, FL setups are vulnerable to byzantine attacks, where malicious clients may poison their local data\cite{datapoison, datapoison2} or local updates\cite{poisonFang, poison2, poison3, poison4}, worsening the performance of the global model.  

This issue is particularly pertinent to multi-agent coordination. For example, a multi-agent system may rely on a vision model trained by FL to identify aspects of their environment. If malicious agents harm the model, the agents will struggle to identify relevant aspects of their environment and accomplish their task. 

\begin{figure}[!t]
\centering
\includegraphics[width=0.5\textwidth]{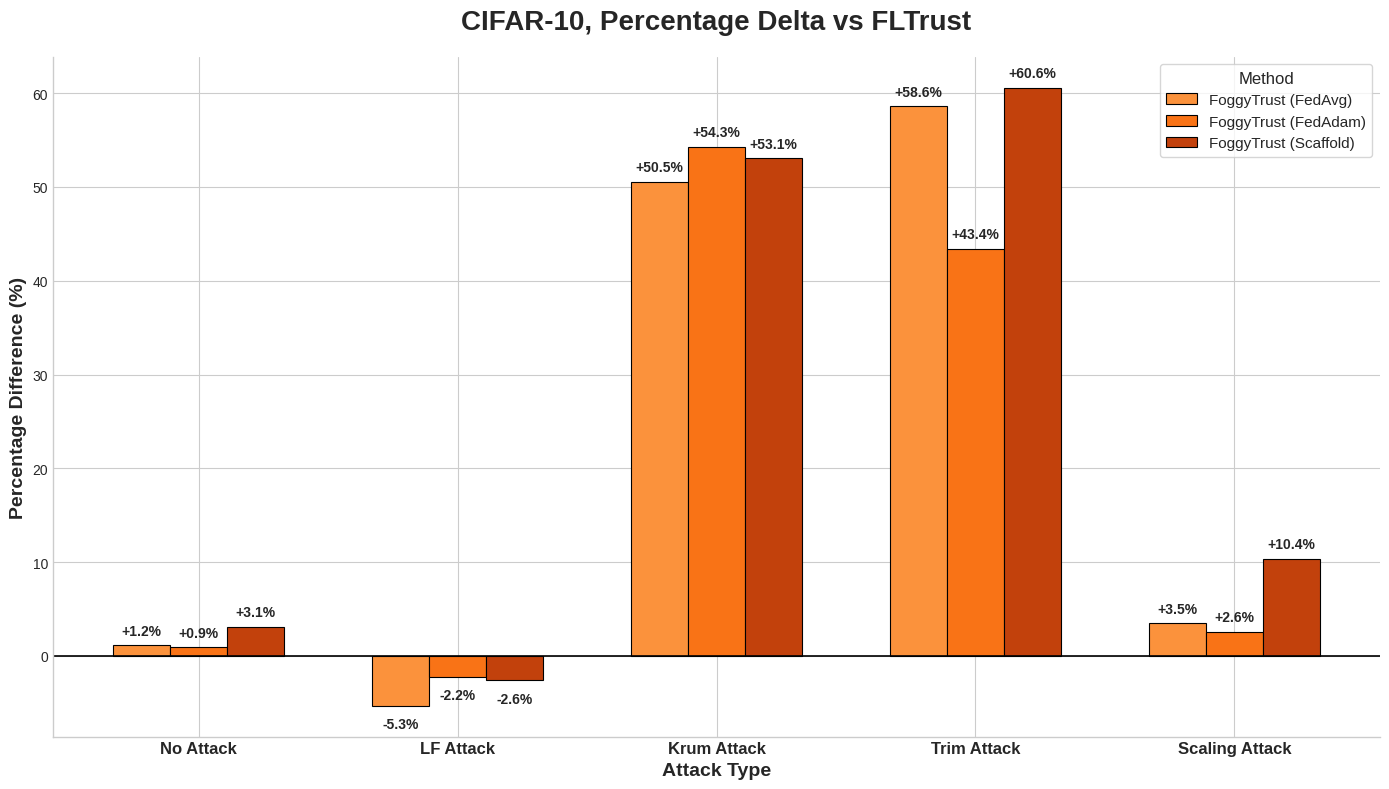}
\caption{Comparing our FoggyTrust algorithms as a percentage of the FLTrust(SOTA baseline) scores on the CIFAR-10 dataset across all attacks. Our method FoggyTrust performs the best for harer attacks like Krum and Trim attack, but has limitations for data poisoning attacks.
}
\label{fig:CIFAR}
\end{figure}

\section{Previous Work}
\subsection{Background on Byzantine-Robust FL Methods}
Byzantine-robust FL methods seek to solve that problem. That is, they attempt to enable FL to successfully occur in situations where clients may be malicious or corrupted. Many of these methods, including Krum\cite{Krum}, Trimmed mean\cite{Mean/Median}, and Median\cite{Mean/Median}, are robust aggregation rules that use statistical techniques to determine if an update is malicious. However, studies have shown that these aggregation rules are still vulnerable to carefully crafted attacks\cite{poisonFang}. These attacks exploit the statistical techniques in a variety of ways.

Fang et al. \cite{poisonFang} created a general framework for these attacks, which adapts to each robust aggregation method. In essence, their framework creates an optimization problem where they attempt to update the global model along the direction most opposite to a non-malicious update. They tested their framework on the Krum and Trimmed mean aggregation methods, and showed that their attacks, the Krum Attack and Trim Attack respectively, severely degraded performance. 

\subsection{FLTrust Algorithm}

Cao et al.\cite{FLTrust} recognized that robust-aggregation methods relying on statistical techniques alone would remain vulnerable to carefully crafted attacks. As a result, they extend previous methods by allowing the service provider to bootstrap trust. By developing a notion of how much the service provider trusts each local update, the service provider can accept trustworthy updates and reject or minimize untrustworthy ones. 

They provide this notion of trust by crafting a small ($<\!100$ samples) root dataset for the service provider, which is unique for FL where data is typically entirely decentralized. The root dataset must be non-malicious and representative of the data seen in the clients. 

The first two steps of an iteration of the algorithm are largely unchanged from typical FL methods. First, they synchronize the models by sending the service provider's global model, $w$ to the clients participating in this round of updates. Secondly, the clients make local updates $g_i$ to the model by fine tuning on the data that is present on their devices. Novelly, they also compute the service provider's local update, $g$, based on the carefully crafted root dataset. 

Next, each clients local update $g_i$ is compared to $g$ by taking the cosine similarity of the two updates. A more aligned update (cosine similarity closer to 1)  is considered more trustworthy, whereas less aligned updates (cosine similarity closer to -1) are considered less trustworthy. By the construction of their aggregation rule, a local update with a negative trust score would still contribute toward the global model, despite them being considered very untrustworthy. Thus, they ReLU clip the trust score to ensure updates with negative cosine similarities aren't included. Finally, they normalize each local update $g_i$ to match the magnitude of $g$, protecting against scaling attacks, where malicious agents attempt to have a disproportionately large effect on the global model\cite{poison3}. Then, they aggregate the normalized local updates by weighing each update $g_i$ proportionally to its ReLU-clipped trust score $TS_i$.

Their defense model successfully defended against the Krum and Trim attack, among others, with some experiments showing no reduction in accuracy in some instances, and only a small reduction in others\cite{FLTrust}. However, their defense relies a strong assumption that weakens its practicality: FLTrust assumes that the data each client possesses is relatively independent and identically distributed (iid).

In reality, the applications that employ FL often have very heterogeneous information. For example, consider the next-word application of FL on Android GBoard\cite{Gboard}. This is trained on language data from billions of different users, which certainly don't speak in the same manner. 

Heterogeneity presents a problem for FLTrust for two reasons. First, the heterogeneity of a dataset can cause its root of trust to not be sufficiently representative of all clients. If we have an unrepresentative root datset, then our trust scores will be brittle. Since the aggregation rule directly depends on our trust scores, it begins to break down the algorithm. For example, FLTrust may deem updates untrustworthy even if they're not malicious, as they sufficiently differ from the root datasets update due to the non-iid-ness of the data. 

Secondly, non-iidness in data in FL is a problem in and of itself. Reddie et al. describes two primary issues FedAvg runs into in certain applications\cite{FedAdam}. First, it can suffer from "client drift," where the local updates begin to deviate from the global model as they move toward locally optimal solutions. Secondly, the method is non-adaptive, making it unable to adapt to client variability. 

Several methods have been created to correct for these circumstances. To address the first issue, SCAFFOLD\cite{Scaffold} introduces client and server control variates that act as correction terms during local training, reducing the bias caused by heterogeneous client data and mitigating client drift. By keeping local updates better aligned with the global objective, SCAFFOLD improves convergence in non-IID settings. To address the second issue, FedAdam\cite{FedAdam} replaces FedAvg’s standard server update with an adaptive optimization method, enabling the server to adjust more effectively to variability in client updates across rounds. These two approaches therefore target complementary weaknesses of FedAvg under heterogeneity: SCAFFOLD reduces drift in local optimization, while FedAdam improves the adaptivity of global aggregation. However, to our knowledge, no heterogeneity-correcting methods that are byzantine-robust exist, and FLTrust makes no effort to resolve these heterogeneity problems within their techniques. 

Thus, global non-iidness presents a substantial issue for FLTrust, but it is extremely prevalent in application. However, we believe that a weaker assumption may be more realistic: that data can live in families of locally homogeneous pockets despite a globally heterogeneous structure. For instance, consider the GBoard example, and assume that it serves all English speakers in the United States. Then, people in Wisconsin may talk very similarly to each other, but differently than people in Alabama. Our primary objective is to exploit \textit{local homogeneity} to outperform FLTrust in highly heterogeneous environments.

Specifically, in this paper, we have three main contributions over the baseline method FLTrust:
\begin{itemize}
\item We employ root datasets across many fog nodes instead of one at the global server, localizing the trust scores to groups of clients that are more homogeneous, and aggregate the updates at that level with the locally computed trust scores.
\item We create a hierarchical nature to fold in updates computed at each fog node, and swap out the 2nd-level aggregator with other federated learning aggregation rules that are more robust to client drift.
\item We test our method against FLTrust in a real-world dataset such as SnapshotSafari\cite{SnapshotSafari}.
\end{itemize}
Our code is publicly accessible at the following link: \href{https://github.com/emmanuel2406/FoggyTrust.git}{https://github.com/emmanuel2406/FoggyTrust.git}

\begin{figure}[!b]
\centering
\includegraphics[width=0.5\textwidth]{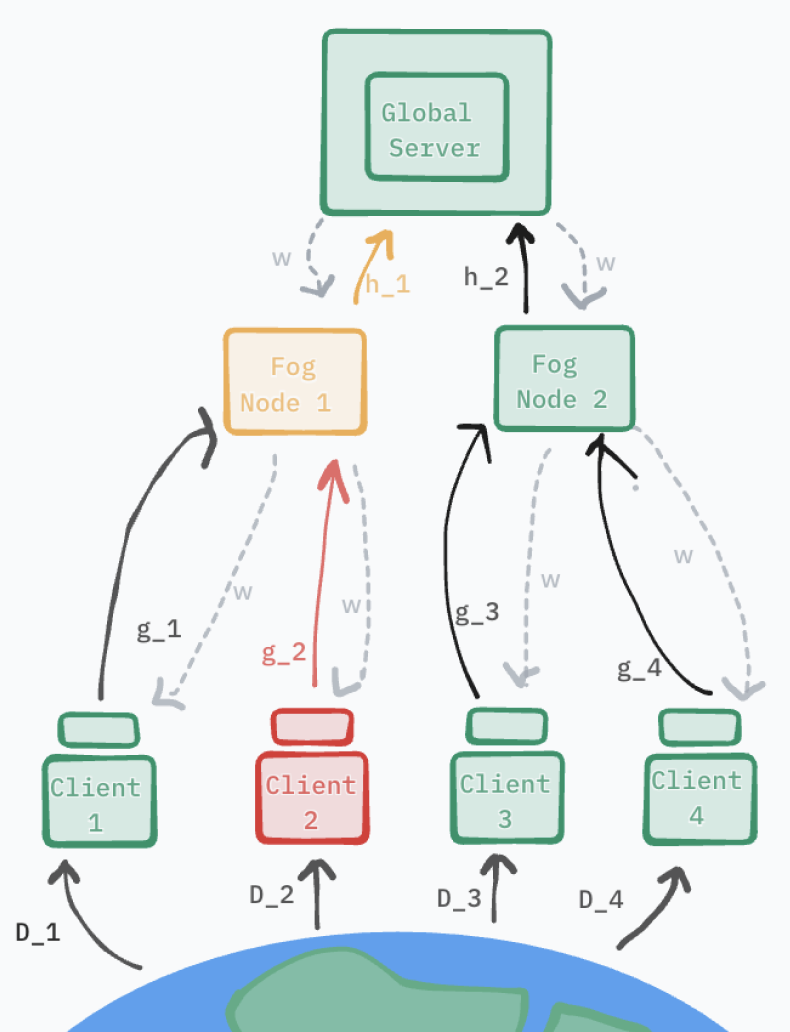}
\caption{High level overview of \texttt{FoggyTrust} with FedAvg, where \textbf{stage i} applies FLTrust between the clients and the fog nodes and \textbf{stage ii} applies FedAvg between the weight updates of each fog node and the global server.}
\label{fig:hfl}
\end{figure}

\section{FoggyTrust}
We propose \textsc{FoggyTrust}, a hierarchical FL method that bootstraps trust and performs well in globally heterogenous applications that possess local homogeneity. See Figure \ref{fig:hfl} for a visual overview.

FoggyTrust extends FLTrust in the following manner. Rather than having a single global server run the FLTrust algorithm for all clients, the service provider instead communicates with fog nodes. The fog nodes serve as local service providers for data that we assume to be relatively iid. Following our previous example, there could be a fog node at each state, where people are assumed to talk relatively similarly. We construct a root dataset for each of these fog nodes, under the assumption that it is relatively representative of its clients data. 

Thus, at each iteration of the algorithm, clients make local updates which they send to their respective fog node rather than the global server. Each fog node aggregates the client data corresponding using FLTrust to have a fog node level local update. Then, each of the fog nodes local updates send their aggregated update to the global server. The global server then uses an aggregation rule (such as FedAvg \cite{FedAvg}) to fold in the updates of each fog node.

FoggyTrust with FedAvg overcomes the heterogeneity between server updates and each client update. Since each local group is assumed to be relatively iid, the root dataset at each fog node can thus be representative of the data within. This ensures that the trust scores are accurate, and the aggregation rule performs as intended at the lower level. The higher level aggregation does not compute trust scores, thus all FLTrust-level issues with heterogeneity are addressed by this approach.

We also seek to address the second problem of heterogeneity, which in our instance would be client drift between fog nodes. Despite each fog node being relatively iid within itself, the local regions are still globally heterogenous. Thus, "client drift" is still a potential issue. However, our hierarchical nature enables us to resolve this using existing aggregators specialized for heterogeneity: SCAFFOLD\cite{Scaffold} mitigates client drift through control variates that correct biased local updates, whereas FedAdam\cite{FedAdam} uses adaptive server-side optimization to improve convergence and accuracy in heterogeneous environments. Accordingly, we modify our second-level aggregation mechanism to better account for heterogeneity and enhance performance.

To our knowledge, FoggyTrust is the first byzantine-robust aggregation method specialized for heterogeneous applications. 

\section{Evaluation}
\subsection{Experimental Setup}
 We aim to test the robustness and generality of three new learning methods:
\begin{enumerate}
    \item \textsc{FoggyTrust(FedAvg)}: Our vanilla FoggyTrust algorithm with FedAvg \cite{FedAvg} used to aggregate at the second level
    \item \textsc{FoggyTrust(FedAdam)}: FedAdam is used to aggregate at the second level \cite{FedAdam}
    \item \textsc{FoggyTrust(SCAFFOLD)}: SCAFFOLD is used to aggregate at the second level \cite{Scaffold}
\end{enumerate}
The exact parameters for each dataset is given in Table \ref{tab:parameters}.
\subsubsection{Datasets}
We test our FoggyTrust on four image classifcation datasets of varying difficulty: MNIST, Fashion-MNIST, CIFAR-10 and Snapshot Safari. Our first step was to replicate the results from FLTrust \cite{FLTrust} by building upon the official implementation by one of the main authors \href{https://minghongfang.com}{https://minghongfang.com}. 

We follow previous work \cite{FLTrust} to partition the dataset across different clients. For MNIST, FashionMNIST and CIFAR-10, each with 10 label classes, we assign each client a biased label so that 50\% of their data points are from that label and the rest are uniformly distributed across the other label classes, so the other groups have probability $\frac{1-0.5}{10-1} \approx 0.056$. This simulates non-IID local training data.

\textbf{MNIST}: MNIST  \cite{MNIST} consists of 10 digits from 0 to 9 with 60,000 training examples

\textbf{Fashion-MNIST}: Fashion-MNIST \cite{Fashion-MNIST} has 60,000 fashion images distributed across 10 categories of different items in the fashion domain

\textbf{CIFAR-10}: CIFAR-10\cite{CIFAR} has RGB color images with 50,000 training examples distributed across 10 semantic categories.

\begin{figure}[!t]
\centering
\includegraphics[width=0.5\textwidth]{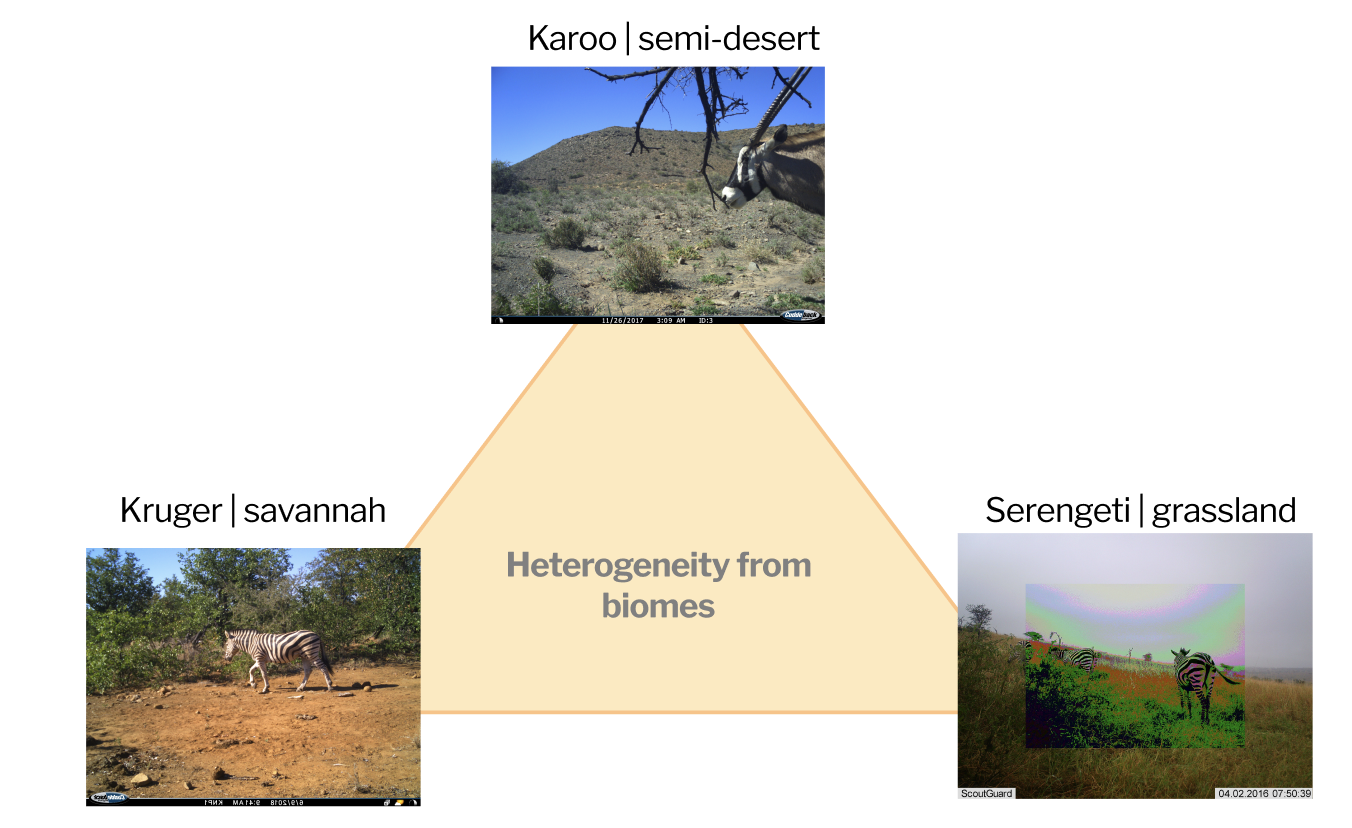}
\caption{Overview of Snapshot Safari setup where we choose three tapped camera environments that are highly heterogenous with respect to each other}
\label{fig:SnapshotSafariSetup}
\end{figure}

\textbf{Snapshot Safari}: Snapshot Safari \cite{SnapshotSafari} is a real-world dataset consisting of 4 Million images from 15 camera tapping projects in the safari. We pick three of these projects which have different biomes, as seen in Figure \ref{fig:SnapshotSafariSetup}. Namely, Karoo in South Africa being a semi-desert, Kruger in South Africa being a savanna, and Serengeti in Tanzania being a grassland. The task here is a multi-label classification task with 151 categories, but without a baseline to ensure proper replication. With limited compute available on this more expansive dataset, we only train for 200 iterations per configuration.

\subsubsection{Global Models}
To show the generality of our method we train different types of models across the different datasets. Particularly, we use a Convolutional Neural Network (CNN) identical to the one used in \cite{FLTrust} for MNIST and Fashion-MNIST. CIFAR-10 and Snapshot Safari are significantly harder tasks so we train a model based on the ResNet-20 architecture \cite{resnet}.

\subsubsection{Poisoning Attacks}
Each attack uses the same parameters as in \cite{FLTrust}.
Label flipping(LF) attack is our only data poisoning attack, which essentially flips $l$ to be $M-l - 1$ for a dataset with $M$ classes.

We also test on Trim, Krum, and Scaling Attack which are local model poisoning attacks. Trim and Krum are both optimized against the Trimmed-Mean and Krum aggregation rules. The Scaling Attack is a targeted attack that scales its local update contribution by $\lambda \gg 1$ before the server handles it. We also measure the attack success rate (lower is better) on top of the usual test accuracy for this attack.

\subsubsection{Extension Setup}
For MNIST, Fashion-MNIST and CIFAR-10, we have 100 clients each and so we group them into client clusters based on their biased label. This gives 10 groups of 10 clients. Each group will be assigned a "fog node" where a local FLTrust algorithm will perform aggregation and then our FoggyTrust hierarchical setup allows an aggregation of each fog node's update vector to the global server.

Analogous to the sampling of the reference dataset for the global server in FLTrust \cite{FLTrust}, we sample each fog node's reference dataset based on the same biased probability used for their client groups.

For Snapshot Safari, we have 3 groups of 10 clients, one for each biome. In this case, the heterogeneity is implicit in the differences in camera tap locations. The fog node reference datasets are similarly sampled.

\subsection{Results}
\subsubsection{MNIST}
As seen in Table \ref{tab:MNIST}, we find FoggyTrust(FedAvg) beats all the other baselines (including FLTrust) under no attack, Trim and Scaling Attack. However, FLTrust still performs best or equal best for the Label Flipping and Krum Attack. FLTrust also has the best attack success rate for the scaling attack. 

The second level aggregator (both FedAdam and SCAFFOLD) does not seem to help final test accuracies. These methods aim to help the second problem addressed which is the client drift between different client groups. The distribution of digit classes may be close enough under MNIST that the overhead of a more complicated aggregation rule outweighs the potential benefits of mitigating this issue. 
\\
\subsubsection{Fashion-MNIST}
We compare our vanilla FoggyTrust with each of the previous defense methods in Table \ref{tab:Fashion-MNIST}. FoggyTrust(FedAvg) beats all the other baselines under Krum and Scaling attack. FLTrust performs quite well under this dataset beating our algorithm under no attack and Trim Attack. 

The data poisoning attack, label flipping attack has the simplest baseline FedAvg perform the best: 0.8943 vs 0.8813 for FLTrust \& 0.8784 for FoggyTrust(FedAvg). 

The margins between the methods are still small enough to partially be attributed to noise in training, and so we also consider harder datasets to see a more clearer separation.
\\
\subsubsection{CIFAR-10}
We compare all three of our methods to FLTrust in Table \ref{tab:CIFAR}. Under all attack setups except the Label Flipping attack, all three of our new methods beat FLTrust. 

Comparing the different versions of FoggyTrust, the harder task provided by the CIFAR-10 dataset shows promising gains by swapping out the FedAvg algorithm at the second level of our hierarchical algorithm. FoggyTrust with FedAdam performs the best under with a score of 0.5501. FoggyTrust with SCAFFOLD gives more consistent gains with the best scores under no attack, Trim Attack and Scaling attack (both attack success rate and testing accuracy). This method also beats FoggyTrust with FedAvg under Krum attack. 

Figure \ref{fig:CIFAR} highlights the magnitude of difference between the method's performance. By measuring it as a percentage delta of the FLTrust baseline, we see that FLTrust does not perform as well for the Krum and Trim attacks which both target the direction of the gradient update vectors. On the contrary, FoggyTrust calculates the trust scores at only a local lever which preserves the robustness of the model's federated learning especially under local model poisoning attacks.

As a disclaimer, our experiments do not replicate the abnormally high accuracy results from \cite{FLTrust} under this dataset. We rationalize the difference from other settings to be due to the  training of a larger model Resnet-20 which may have had an undisclosed initialization or checkpoint, or the dataset may have had some prepocessing step as a necessary transformation.

\subsubsection{Snapshot Safari}

\begin{figure}[!ht]
\centering
\includegraphics[width=0.5\textwidth]{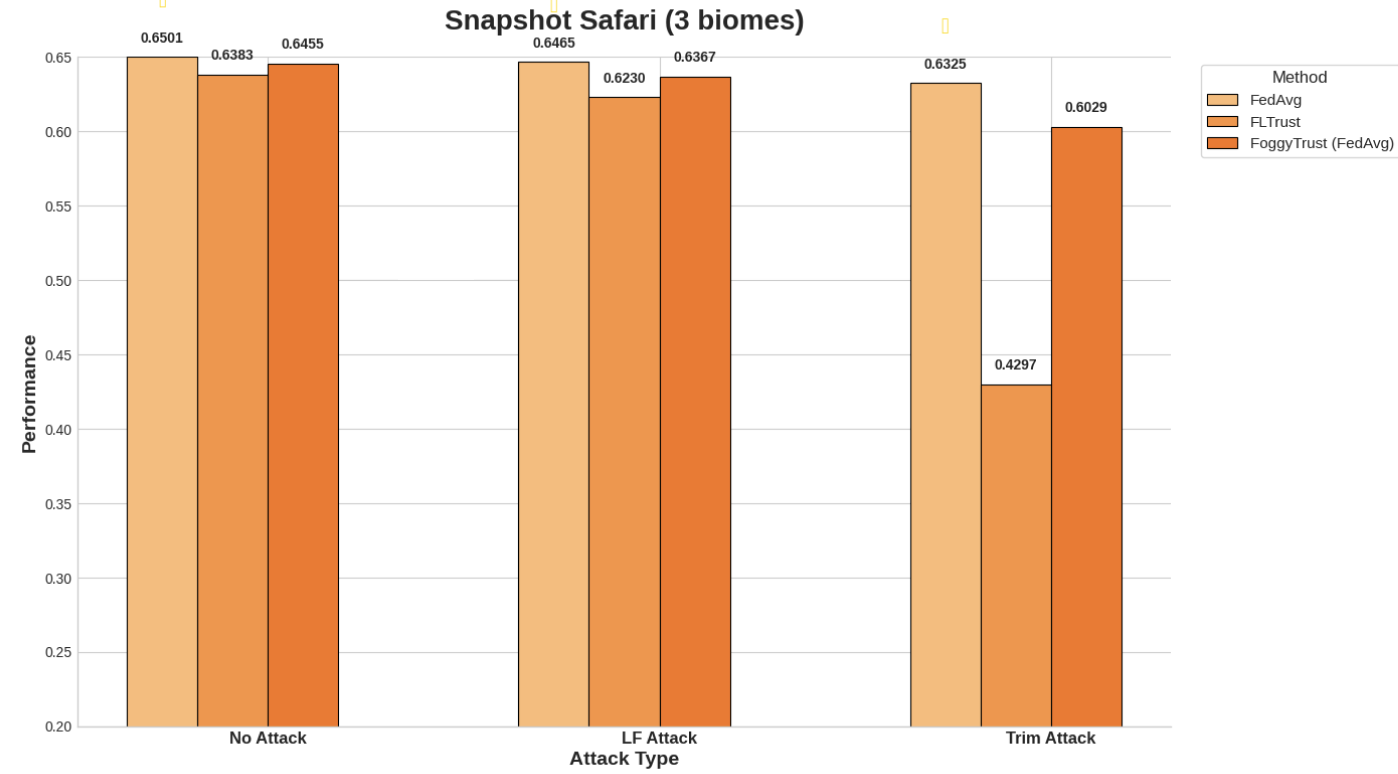}
\caption{Comparing our FoggyTrust algorithms as a percentage of the FLTrust scores on our subset of the Snapshot Safari dataset across 3 attack settings}
\label{fig:Snapshotsafari}
\end{figure}

The full results are shown in Figure \ref{fig:Snapshotsafari}. We only test on No attack, Label Flipping attack and Trim Attack. The simplest algorithm, FedAvg, performs the best under all three settings. For the two attack settings, this seems counter-intuitive as, this method should be the most susceptible to attacks. 

Our FoggyTrust algorithm does offer some value by bridging the gap in performance between FedAvg and FLTust. This is especially true for Trim Attack, where FLTrust has a drop in their test accuracy from 0.6383 to 0.4297, while FoggyTrust has a more graceful drop from 0.6455 to 0.6029.

\subsection{Partitioned FoggyTrust}
 By default, we replicate the size of the reference datasets given to each client group. We also tested a partitioned version of our FoggyTrust algorithm, where we instead \textit{partition} the server root dataset equally across all client groups. So for the datasets with 10 fog nodes, each one will have a reference dataset of 10 data points, instead of 100. This leads to slower convergence compared to our original version with accuracy scores of 0.7882 (vs 0.9627) under no attack and 0.7840 (vs 0.9604) under scaling attack for MNIST. \cite{FLTrust} has tuned the dataset size to be 100, and so we adopt this size length even though it would mean a higher overhead for the model administrator to collect data as we scale the number of client groups.

\section{Discussion and Limitations}
\subsection{FoggyTrust demonstrates improved robustness under stronger attacks and more challenging datasets}

The most significant performance gains of FoggyTrust over the FLTrust baseline are observed on the CIFAR-10 dataset under both Trim and Krum attacks. Notably, these attacks are specifically designed to manipulate the direction of client update vectors. In scenarios where the server-side reference dataset is not representative of the heterogeneous client distributions, FLTrust may assign low trust scores to benign client updates while inadvertently allowing adversarial updates to influence the global aggregation.

In contrast, the hierarchical structure of FoggyTrust yields more reliable trust estimation by leveraging local homogeneity within each fog node. By computing trust scores at the level of relatively homogeneous client groups, the method mitigates the mismatch between the reference dataset and client distributions that can degrade FLTrust’s effectiveness.

Although the grouping strategy in synthetic settings such as CIFAR-10 explicitly enforces homogeneity, similar structure often emerges in real-world federated learning deployments. In particular, client data distributions are frequently correlated with latent factors such as geographic location, leading to naturally clusterable subpopulations. This observation suggests that the assumptions underlying FoggyTrust are not only practical but also aligned with real-world data heterogeneity patterns.

\subsection{Modular Advantage of Two-Level Aggregation}
One of the benefits that comes with having two levels of aggregations is that we can swap in any aggregation rule between the fog node  and the global server level. While FedAvg should do well, we note that it suffers from client drift and unadaptability, which will be exacerbated at the second level where client groups are heterogeneous with respect to each other. 

From the CIFAR-10 experiment, we see that SCAFFOLD \cite{Scaffold} consistently had gains over our vanilla implementation of FedAvg at the second level. SCAFFOLD introduces control variates specifically to correct the client drift. For Krum, Trim and Scaling attacks, this makes updates less sensitive to local deviations by aligning each update specifically with a global objective. This reduction in variance across fog node groups allows for a more robust learning process that makes it tougher for malicious clients to skew the gradient direction of the global model. 

SCAFFOLD is just one possible aggregation rule. We also tested FedAdam \cite{FedAdam} which focused more on the global aggregation rule than the local optimization and so it was not as robust for all attack vectors. Our paper hopes to open up the possibilities for future trust network architectures which potentially synergize better with the trust score calculations on level one. 

\subsection{Limitations for Data Poisoning Attacks}
Although our hierarchical extension proves to be valuable for local model poisoning attacks, the gains do not translate over to data poisoning attacks. Specifically, the label flipping attack for the Fashion-MNIST and CIFAR-10 datasets. 

The main reason for this shortfall is due to the silent nature of the label-flipping attacks. The resulting gradients for training on the wrong labels are still mathematically plausible and consistent with one another. For trust score calculation, if more of the malicious clients updates are higher weighted than the honest updates, this will degrade the accuracy of the model being trained. 

Because the trust scores are calculated for 10 clients in the case of FoggyTrust (within every client group) compared to 100 in the case of FLTrust, the room for error is a lot smaller. This is coupled with the notion that updates will have higher variance for more complicated data distributions such as those sampled from Fashion-MNIST and CIFAR-10. Future work should investigate the scaling laws as we increase the number of clients to determine the optimal ratio of clients per fog node.

\subsection{Deployment in the Wild}

While the majority of our experiments are conducted on synthetic benchmarks, we additionally evaluate the practicality of FoggyTrust on a real-world dataset, Snapshot Safari \cite{SnapshotSafari}. Previous work suggests that fog nodes may  offer communication and scalability advantages over flat federated learning systems. However, a detailed analysis of these systems-level considerations is beyond the scope of this work.

FoggyTrust retains a key advantage over FLTrust in this setting: inter-group heterogeneity (e.g., across biomes) does not adversely affect trust score estimation, as trust is computed within locally homogeneous fog nodes. In contrast, FLTrust relies on a single global reference dataset, making it more susceptible to distributional mismatch across highly heterogeneous environments.

That said, the Snapshot Safari dataset introduces substantial intra-group variability, as data collected within a single biome remains inherently unstandardized. Variations in environmental conditions can induce significant local heterogeneity which include weather, illumination (e.g., day versus night), and camera positioning. These factors may degrade the reliability of trust estimation even within fog nodes.

As a result, the benefits of trust-based aggregation become less pronounced in this setting. In particular, the additional computational and statistical overhead associated with trust score estimation may outweigh its robustness advantages, with simpler methods such as FedAvg exhibiting competitive or superior performance.

\section{Conclusion}

In this work, we introduced FoggyTrust, a hierarchical trust-based aggregation framework for federated learning in heterogeneous environments. Our results demonstrate that FoggyTrust consistently outperforms FLTrust on more challenging datasets and under adversarial settings, particularly when the assumption of a globally representative reference dataset is violated. By localizing trust computation within relatively homogeneous client groups, FoggyTrust improves the reliability of trust scores and enhances robustness against attacks that manipulate update directions.

A key takeaway from this work is that hierarchical design, despite introducing additional system complexity, can yield substantial benefits in federated learning. In particular, decoupling local trust estimation from global aggregation enables modularity: different aggregation strategies (e.g. SCAFFOLD) can be incorporated at higher levels to address phenomena such as client drift. This highlights the potential of combining robustness mechanisms with heterogeneity-aware optimization techniques.

We further evaluated FoggyTrust on the Snapshot Safari dataset, demonstrating its applicability in real-world, highly heterogeneous settings. While results suggest that trust-based methods may incur additional overhead in such environments, FoggyTrust narrows the performance gap between robust and non-robust methods and provides a principled framework for handling structured heterogeneity. These findings indicate potential for deployment in safety-critical applications, such as distributed sensor networks or wildlife monitoring systems, where resilience to corrupted or adversarial inputs is essential.

Several directions remain for future exploration. First, a more detailed systems-level analysis of FoggyTrust is needed, particularly with respect to communication efficiency, scalability, and latency in real-world deployments. Second, improving robustness to data poisoning attacks remains an open challenge. Third, adaptive or learned grouping strategies could replace manually defined fog nodes, enabling the system to dynamically discover latent structure in client distributions. Finally, integrating more advanced second-level aggregation methods or jointly optimizing trust estimation and aggregation may further enhance performance in highly heterogeneous and adversarial environments.

\section{Acknowledgements}
We would like to express our sincere gratitude to Dr.\ Stephanie Gil and Karen Li for their invaluable feedback and guidance throughout this project. Their insights and support greatly contributed to the development and refinement of this work.

\begin{table*}[t]
\centering
\caption{Default parameters used for FL experiments}
\label{tab:parameters}
\resizebox{\textwidth}{!}{%
\begin{tabular}{|c|l|c|c|c|c|}
\hline
 & \textbf{Explanation} & \textbf{MNIST} & \textbf{Fashion-MNIST} & \textbf{CIFAR-10} & \textbf{Snapshot Safari} \\
\hline
$n$ & \# clients & \multicolumn{3}{c|}{100} & 30 \\
\hline
$\tau$ & \# clients selected in each iteration & \multicolumn{4}{c|}{$n$} \\
\hline
$R_l$ & \# local iterations & \multicolumn{4}{c|}{1} \\
\hline
$R_g$ & \# global iterations & 2000 & 2500 & 1500 & 200 \\
\hline
$b$ & batch size & \multicolumn{2}{c|}{32} & \multicolumn{2}{c|}{64} \\
\hline
$\alpha \cdot \beta$ & combined learning rate & $3 \times 10^{-4}$ & $6 \times 10^{-3}$ & $2 \times 10^{-4}$ & $1.5 \times 10^{-4}$ \\
\hline
$\frac{m}{n}$ & fraction of malicious clients (\%) & \multicolumn{4}{c|}{20} \\
\hline
$m$ & \# malicious clients & \multicolumn{3}{c|}{20} & 6 \\
\hline
$f$ & Krum parameter & \multicolumn{4}{c|}{$m$} \\
\hline
$k$ & Trim-mean parameter & \multicolumn{4}{c|}{$m$} \\
\hline
$|D_0|$ & size of each root dataset & \multicolumn{4}{c|}{100} \\
\hline
\end{tabular}%
}
\end{table*}

\begin{table*}[htbp]
\centering

\caption{The testing accuracy scores of different FL methods
under different attacks and the attack success rates of the
Scaling attacks for the \textbf{MNIST} dataset. The results for the Scaling attacks format the attack success rate of the targeted attack below each test accuracy. *our contributions
}
\label{tab:MNIST}
\begin{tabular}{lcccccccc}
\hline
 & FedAvg & Krum & Trim-Mean & Median & FLTrust 
 & \makecell{*FoggyTrust\\(FedAvg)} 
 & \makecell{*FoggyTrust\\(FedAdam)} 
 & \makecell{*FoggyTrust\\(SCAFFOLD)} \\
\hline
No Attack      
& 0.9550 & 0.8966 & 0.9400 & 0.9385 & 0.9522 
& \textbf{0.9627} & 0.9549 & \textbf{0.9626} \\

LF Attack      
& 0.9377 & 0.9007 & 0.8915 & 0.9412 & \textbf{0.9492} 
& 0.9480 & 0.9520 & \textbf{0.9491} \\

Krum Attack    
& 0.8921 & 0.0983 & 0.7979 & 0.1083 & \textbf{0.9391} 
& 0.9366 & 0.9362 & 0.9290 \\

Trim Attack    
& 0.7489 & 0.8986 & 0.7043 & 0.6780 & 0.9199
& \textbf{0.9531} & 0.9453 & 0.9479 \\

Scaling Attack 
& \makecell{0.0980 \\ {\scriptsize 1.0000}} 
& \makecell{0.8855 \\ {\scriptsize 0.0146}} 
& \makecell{0.6677 \\ {\scriptsize 0.0340}} 
& \makecell{0.8788 \\ {\scriptsize 0.0095}} 
& \makecell{0.9386 \\ {\scriptsize \textbf{0.0038}}} 
& \makecell{\textbf{0.9604} \\ {\scriptsize 0.0058}} 
& \makecell{0.9465 \\ {\scriptsize 0.0068}} 
& \makecell{0.9580 \\ {\scriptsize 0.0049}} \\
\hline
\end{tabular}
\end{table*}

\begin{table*}[htbp]
\centering

\caption{The testing accuracy scores of different FL methods
under different attacks and the attack success rates of the
Scaling attacks for the \textbf{Fashion-MNIST} dataset. The results for the Scaling attacks format the attack success rate of the targeted attack below each test accuracy. *our contributions
}
\label{tab:Fashion-MNIST}
\begin{tabular}{lcccccc}
\hline
 & FedAvg & Krum & Trim-Mean & Median & FLTrust & \makecell{*FoggyTrust\\(FedAvg)} \\
\hline
No Attack      
& 0.8894 & 0.8198 & 0.8650 & 0.8578 & \textbf{0.8898}& 0.8888 \\

LF Attack      
& \textbf{0.8943} & 0.8408 & 0.7670 & 0.8326 & 0.8813 & 0.8784 \\

Krum Attack    
& 0.1000 & 0.1000 & 0.6961 & 0.7121 & 0.8655 & \textbf{0.8793} \\

Trim Attack    
& 0.1000 & 0.8389 & 0.7218 & 0.7267 & \textbf{0.8952} & 0.8791 \\

Scaling Attack 
& \makecell{0.1000 \\ {\scriptsize 1.000}} 
& \makecell{0.8378 \\ {\scriptsize 1.000}} 
& \makecell{0.1000 \\ {\scriptsize 1.000}} 
& \makecell{0.1000 \\ {\scriptsize 1.000}} 
& \makecell{0.8865 \\ {\scriptsize \textbf{0.000}}} 
& \makecell{\textbf{0.8952} \\ {\scriptsize \textbf{0.000}}} \\
\hline
\end{tabular}

\end{table*}

\begin{table*}[htbp]
\centering

\caption{The testing accuracy scores of different FL methods
under different attacks and the attack success rates of the
Scaling attacks for the \textbf{CIFAR-10} dataset. The results for the Scaling attacks format the attack success rate of the targeted attack below each test accuracy. *our contributions
}
\label{tab:CIFAR}
\begin{tabular}{lcccc}
\hline
 & FLTrust & \makecell{*FoggyTrust\\(FedAvg)} & \makecell{*FoggyTrust\\(FedAdam)} & \makecell{*FoggyTrust\\(SCAFFOLD)} \\
\hline
No Attack      
& 0.5334 & 0.5396 & 0.5384 & \textbf{0.5501} \\

LF Attack      
& \textbf{0.5281} & 0.5001 & 0.5165 & 0.5143 \\

Krum Attack    
& 0.3213 & 0.4837 & \textbf{0.4957} & 0.4918 \\

Trim Attack    
& 0.3257 & 0.5165 & 0.4671 & \textbf{0.5230} \\

Scaling Attack 
& \makecell{0.5036 \\ {\scriptsize 0.0060}} 
& \makecell{0.5211 \\ {\scriptsize 0.0371}} 
& \makecell{0.5165 \\ {\scriptsize 0.0060}} 
& \makecell{\textbf{0.5559} \\ {\scriptsize \textbf{0.0010}}} \\
\hline
\end{tabular}

\end{table*}

\FloatBarrier
\raggedbottom
\bibliographystyle{IEEEtran}
\bibliography{main}

\end{document}